%% file: paper.tex
\documentclass[runningheads]{llncs}

\usepackage{graphicx}
\graphicspath{ {Figures/} }

\usepackage[utf8]{inputenc} 
\usepackage[T1]{fontenc}    
\usepackage{url}            
\usepackage{booktabs}       
\usepackage{amsmath}
\usepackage{amssymb}
\usepackage{amsfonts}       
\usepackage{nicefrac}       
\usepackage{array}
\usepackage[colorinlistoftodos]{todonotes}
\usepackage{xspace}
\usepackage{paralist}
\usepackage{comment}
\usepackage{caption,subcaption}

\newcommand{\param}{\ensuremath{\vct \theta}}

\newcommand{\vct}[1]{\ensuremath{\mathbf{#1}}}

\newcommand{\set}[1]{\ensuremath{\mathcal{#1}}}
\newcommand{\con}[1]{\ensuremath{\mathsf{#1}}}

\newcommand{\T}{\ensuremath{\top}}
\newcommand{\mycomment}[1]{\textcolor{red}{#1}}

\newcommand{\argmin}{\operatornamewithlimits{\arg\,\min}}

\newcommand{\myparagraph}[1]{\smallskip \noindent \textbf{#1}}
\newcommand{\ie}{{i.e.}\xspace}
\newcommand{\eg}{{e.g.}\xspace}

\usepackage{multirow}
\usepackage[normalem]{ulem}
\useunder{\uline}{\ul}{}

\usepackage{algorithm}
\usepackage[noend]{algpseudocode}

\begin{document}

\title{Poisoning Attacks on Algorithmic Fairness}

\author{David Solans\inst{1}\orcidID{0000-0001-6979-9330} \and
Battista Biggio\inst{2}\orcidID{0000-0001-7752-509X	} \and
Carlos Castillo\inst{3}\orcidID{0000-0003-4544-0416}}
\authorrunning{D. Solans et al.}
\institute{Universitat Pomepu Fabra, Barcelona, Spain\\
\email{\{firstname, lastname\}@upf.edu}\\ \and
Università degli Studi di Cagliari, Cagliari, Italy\\
\email{\{firstname,lastname\}@unica.it}\\ \and
Universitat Pomepu Fabra, Barcelona, Spain\\
\email{chato@acm.org}}

\maketitle

\input{00-abstract}
\input{01-introduction}
\input{02-poisoning_fairness}
\input{03-experiments}
\input{04-related_work}
\input{05-conclusions}

\input{06-acknowledgements}

\bibliographystyle{splncs04}
\bibliography{references,bibDB}

\end{document}

%% file: 00-abstract.tex
\begin{abstract}

Research in adversarial machine learning has shown how the performance of machine learning models can be seriously compromised by injecting even a small fraction of poisoning points into the training data. While the effects on model accuracy of such poisoning attacks have been widely studied, their potential effects on other model performance metrics remain to be evaluated. In this work, we introduce an optimization framework for poisoning attacks against algorithmic fairness, and develop a gradient-based poisoning attack aimed at introducing classification disparities among different groups in the data.
We empirically show that our attack is effective not only in the white-box setting, in which the attacker has full access to the target model, but also in a more challenging black-box scenario in which the attacks are optimized against a substitute model and then transferred to the target model.
We believe that our findings pave the way towards the definition of an entirely novel set of adversarial attacks targeting algorithmic fairness in different scenarios, and that investigating such vulnerabilities will help design more robust algorithms and countermeasures in the future.

\keywords{algorithmic discrimination \and algorithmic fairness \and poisoning attacks \and adversarial machine learning \and  machine learning security}
\end{abstract}

%% file: 01-introduction.tex
\section{Introduction}
\emph{Algoritmic Fairness} is an emerging concern in computing science that started within the data mining community but has extended into other fields including machine learning, information retrieval, and theory of algorithms~\cite{hajian2016algorithmic}.
It deals with the design of algorithms and decision support systems that are non-discriminatory, i.e., that do not introduce an unjustified disadvantage for members of a group, and particularly that do not further place at a disadvantage members of an already disadvantaged social group.
In machine learning, the problem that has been most studied to date is supervised classification, in which algorithmic fairness methods have been mostly proposed to fulfill criteria related to parity (equality)~\cite{zafar2017parity}.
Most of the methods proposed 
to date assume benevolence from the part of the data scientist or developer creating the classification model: she is envisioned as an actor trying to eliminate or reduce potential discrimination in her model.

The problem arises when dealing with malicious actors that can tamper with the model development, for instance by tampering with training data.
Traditionally, \emph{poisoning attacks} have been studied in \emph{Adversarial Machine Learning}. These attacks are usually crafted with the purpose of increasing the misclassification rate in a machine learning model, either for certain samples or in an indiscriminate basis, and have been widely demonstrated in adversarial settings (see, e.g., \cite{biggio18}).

In this work, we show that an attacker may be able to introduce algorithmic discrimination by developing a novel poisoning attack. The purpose of this attacker is to create or increase a disadvantage against a specific group of individuals or samples. For that, we explore how analogous techniques can be used to compromise a machine learning model, not to drive its accuracy down, but with the purpose of adding algorithmic discrimination, or exaggerating it if it already exists. In other words, the purpose of the attacker will be to create or increase a disadvantage against a specific group of individuals or samples.

\myparagraph{Motivation.} The main goal of this paper is to show the potential harm that an attacker can cause in a machine learning system if the attacker can manipulate its training data.
For instance, the developer of a criminal recidivism prediction tool~\cite{ProPublica_article} could sample training data in a discriminatory manner to bias the tool against a certain group of people.
Similar harms can occur when training data is collected from public sources, such as online surveys that cannot be fully trusted.
A minority of ill-intentioned users could \emph{poison} this data to introduce defects in the machine learning system created from it.
In addition to these examples, there is the unintentional setting, where inequities are introduced in the machine learning model as an undesired effect of the data collection or data labeling.
For instance, human annotators could systematically make mistakes when assigning labels to images of people of a certain skin color~\cite{buolamwini2018}.

The methods we describe on this paper could be used to model the potential harm to a machine learning system in the worst-case scenario, demonstrating the undesired effects that a very limited amount of wrongly labeled samples can cause, even if created in an unwanted manner.

\myparagraph{Contributions.} This work first introduces a novel optimization framework to craft poisoning samples that against algorithmic fairness. 
After this, we perform experiments in two scenarios: a ``black-box'' attack in which the attacker only has access to a set of data sampled from the same distribution as the original training data, but not the model nor the original training set, and a ``white-box'' scenario in which the attacker has full access to both.
The effects of these attacks are measured using impact quantification metrics.
The experiments show that by carefully perturbing a limited amount of training examples, an skilled attacker has the possibility of introducing different types of inequities for certain groups of individuals.
This, can be done without large effects on the overall accuracy of the system, which makes these attacks harder to detect.
To facilitate the reproducibility of the obtained results, the code generated for the experiments has been published in an open-source repository. \footnote{https://github.com/dsolanno/Poisoning-Attacks-on-Algorithmic-Fairness}.

\myparagraph{Paper structure.}
The rest of this paper is organized as follows. 
Section~\ref{sec:algorithm}, describes the proposed methodology to craft poisoning attacks for algorithmic fairness.
Section~\ref{sec:experiments} demonstrates empirically the feasibility of the new types of attacks on both synthetic and real-world data, under different scenarios depending on the attacker knowledge about the system.
Section~\ref{sec:related-work} provides further references to related work.
Section~\ref{sec:conclusions} presents our conclusions.

%% file: 02-poisoning_fairness.tex
\section{Poisoning Fairness}\label{sec:algorithm}

In this section we present a novel gradient-based poisoning attack, crafted with the purpose of compromising algorithmic fairness, ideally without significantly degrading accuracy.

\myparagraph{Notation.} Feature and label spaces are denoted in the following with $\set X \subseteq \mathbb R^{\con d}$ and $\set Y \in \{-1,1\}$, respectively, with $\con d$ being the dimensionality of the feature space. 
We assume that the attacker is able to collect some training and validation data sets that will be used to craft the attack. We denote them as $\set D_{tr}$ and $\set D_{val}$.
Note that these sets include samples along with their labels.
$L(\set D_{\rm val}, \vct \theta)$ is used to denote the validation loss incurred by the classifier $f_{\vct \theta} : \set X \rightarrow \set Y $, parametrized by $\vct \theta$, on the validation set $\set D_{\rm val}$.
$\set L(\set D_{\rm tr}, \vct \theta)$ is used to represent the regularized loss optimized by the classifier during training.

\subsection{Attack Formulation} 

Using the aforementioned notation, we can formulate the optimal poisoning strategy in terms of the following bilevel optimization:
\begin{eqnarray}
    \label{eq:obj-pois}
    \label{eq:poisoning_problem_outer}
    \max_{\vct x_c} & &  \set A (\vct x_c, y_c) = L( \set D_{\rm val}, \param^\star) \, , \\
    \label{eq:poisoning_problem_inner}
    {\rm s.t.} & &  \param^\star \in  \argmin_{\param} \set L (\set D_{\rm tr} \cup (\vct x_c, y_c), \vct \param) \, , \\
    \label{eq:poisoning_problem_box}
    & & \vct x_{\rm lb} \preceq \vct x_c \preceq \vct x_{\rm ub} \, . 
\end{eqnarray}
The goal of this attack is to maximize a loss function on a set of untainted (validation) samples, by optimizing the  poisoning sample $\vct x_c$, as stated in the outer optimization problem (Eq.~\ref{eq:poisoning_problem_outer}).
To this end, the poisoning sample is labeled as $y_c$ and added to the training set $\set D_{\rm tr}$ used to learn the classifier in the inner optimization problem (Eq.~\ref{eq:poisoning_problem_inner}).
As one may note, the classifier $\vct \param^\star$ is learned on the poisoned training data, and then used to compute the outer validation loss. This highlights that there is an implicit dependency of the outer loss on the poisoning point $\vct x_c$ via the optimal parameters $\vct \param^\star$ of the trained classifier. In other words, we can express the optimal parameters $\vct \param^\star$ as a function of $\vct x_c$, \ie, $\vct \param^\star(\vct x_c)$. This relationship tells us how the classifier parameters change when the poisoning point $\vct x_c$ is perturbed. Characterizing and being able to manipulate this behavior is the key idea behind poisoning attacks.

Within this formulation, additional constraints on the feature representation of the poisoning sample can also be enforced, to make the attack samples stealthier or more difficult to detect. 
In this work we only consider a box constraint that requires the feature values of $\vct x_c$ to lie within some lower and upper bounds (in Eq.~\ref{eq:poisoning_problem_box}, the operator $\preceq$ enforces the constraint for each value of the feature vectors involved). 
This constraint allows us to craft poisoning samples that lie within the feature values observed in the training set.
Additional constraints can be additionally considered, \eg, constraints imposing a maximum distance from an initial location or from samples of the same class, we leave their investigation to future work.
Our goal here is to evaluate the extent to which a poisoning attack which is only barely constrained can compromise algorithmic fairness.

The bilevel optimization considered here optimizes one poisoning point at a time. To optimize multiple points, one may inject a set of properly-initialized attack points into the training set, and then iteratively optimize them one at a time.
Proceeding on a greedy fashion, one can add and optimize one point at a time, sequentially. This strategy is typically faster but suboptimal (as each point is only optimized once, and may become suboptimal after injection and optimization of the subsequent points).

\myparagraph{Attacking Algorithmic Fairness.}
We now define an objective function $\set A(\vct x_c, y_c)$ in terms of a validation loss $L(\set D_{\rm val}, \vct \param)$ that will allow us to compromise algorithmic fairness without significantly affecting classification accuracy.
To this end, we consider the \emph{disparate impact} criterion~\cite{barocas2016big}. This criterion assumes data items, typically representing individuals, can be divided into unprivileged (e.g., people with a disability) and privileged (e.g., people without a disability), and that there is a positive outcome (e.g., being selected for a scholarship).

Although one might argue that there are several algorithmic fairness definitions \cite{narayanan2018translation} that could be used for this analysis, we selected this criterion for its particularity of being incorporated in legal texts in certain countries \cite{feldman2015,zafar2015fairness}. Apart of that, recent studies \cite{friedler2018comparative} show how fairness metrics are correlated in three clusters what means that targeting this criterion will also affect a set of other metrics with similar strength. In addition to this, authors of \cite{Barocas_NIPS_2017} used this metric to illustrate the first of the three historical fairness goals that have been used to define fairness metrics.
Disparate impact is observed when the fraction of unprivileged people obtaining the positive outcome is much lower the fraction of privileged people obtaining the positive outcome.
Formally, to avoid disparate impact:
\begin{equation}
D = \frac{P(\hat Y=1 | G=u)}{P(\hat Y = 1 | G=p} \geq 1-\epsilon\, ,
\end{equation}
where $\hat Y$ is the predicted label, and $G=\{u,p\}$ a \emph{protected attribute} denoting the group of unprivileged ($u$) and privileged ($p$) samples within a set $\set D$.
Disparate impact thus measures the ratio between the fractions of unprivileged and privileged samples that are assigned to the positive class.
Typically, one sets $\epsilon \approx 0.2$ which suggests $D \geq 0.8$ for a fair classifier, as stated by the four-fifths rule of maximum acceptable disparate impact proposed by the US Equal Employment Opportunity Commission (EEOC)~\cite{feldman2015,zafar2015fairness}.
Thus, in general, we should have $D$ values closer to one to improve fairness.

For our poisoning attack to work, we aim to minimize such a ratio, \ie, decreasing the fraction of unprivileged samples for which $\hat y=1$, while increasing the fraction of privileged users which are assigned $\hat y=1$.
For numerical convenience, we choose to maximize the difference (instead of the ratio) between the mean loss computed on the unprivileged and the privileged samples:
\begin{equation}
    L( \set D_{\rm val}, \param) = \underbrace{\sum_{k=1}^p \ell (\vct x_k, y_k, \param)}_{\rm unprivileged} + \lambda \underbrace{\sum_{j=1}^m \ell (\vct x_j, y_j, \param)}_{\rm privileged} \, .
    \label{eq:attack_loss}
\end{equation}
Note that the parameter $\lambda$ here is set to $p/m$ to balance the class priors (rather than dividing the first term by $p$ and the second by $m$). 

To minimize $D$, we would like to have unprivileged samples classified as negative (lower numerator) and privileged classified as positive (higher denominator). 
As we aim to maximize $L( \set D_{\rm val}, \param)$,
we can label the unprivileged samples as positive ($y_k=1$), and the privileged samples as negative ($y_j=-1$). Maximizing this loss will enforce the attack to increase the number of unprivileged samples classified as negative and of privileged samples classified as positive. 

\begin{figure}[t]
  \centering
      \includegraphics[height=0.38\textwidth]{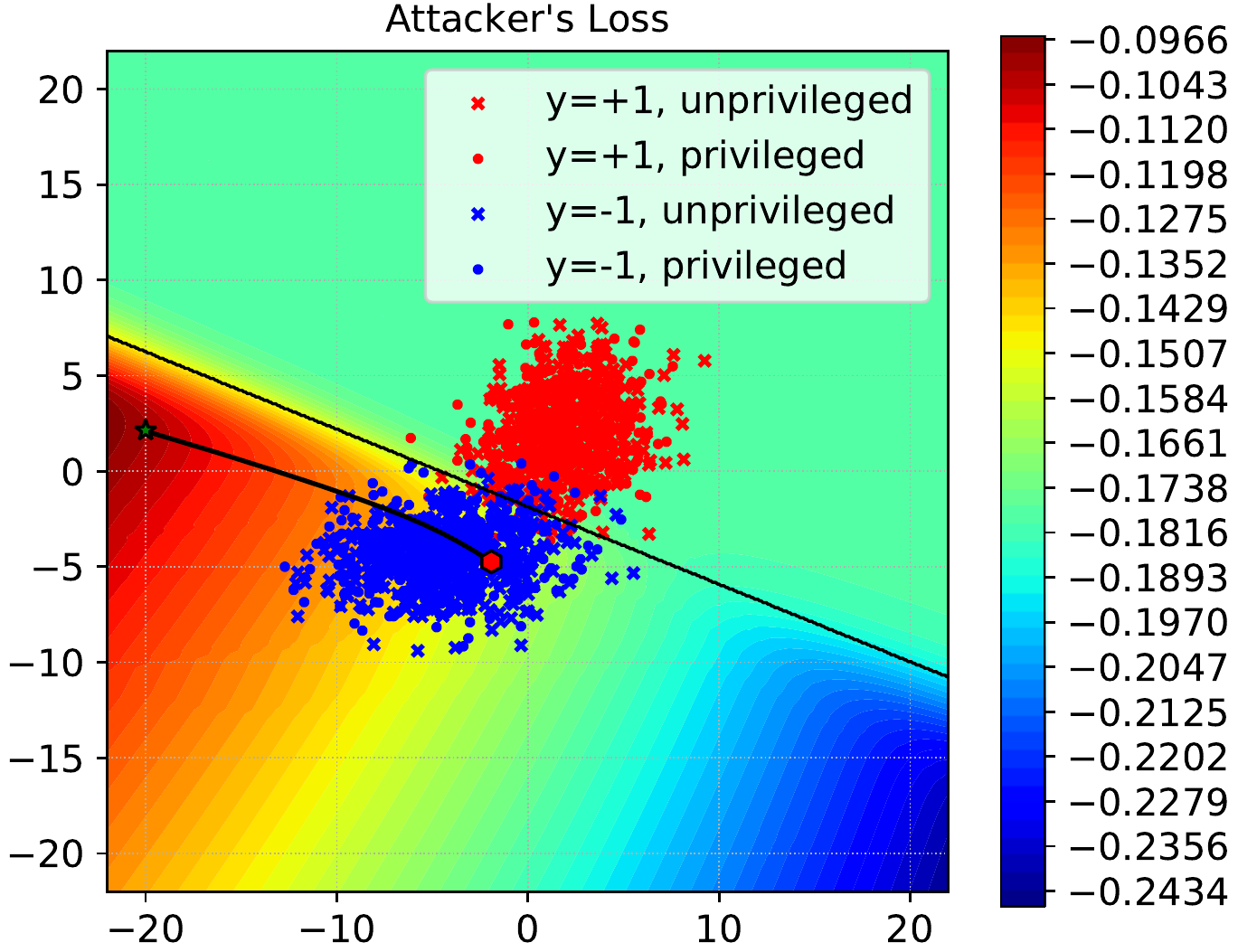}
      \includegraphics[height=0.38\textwidth]{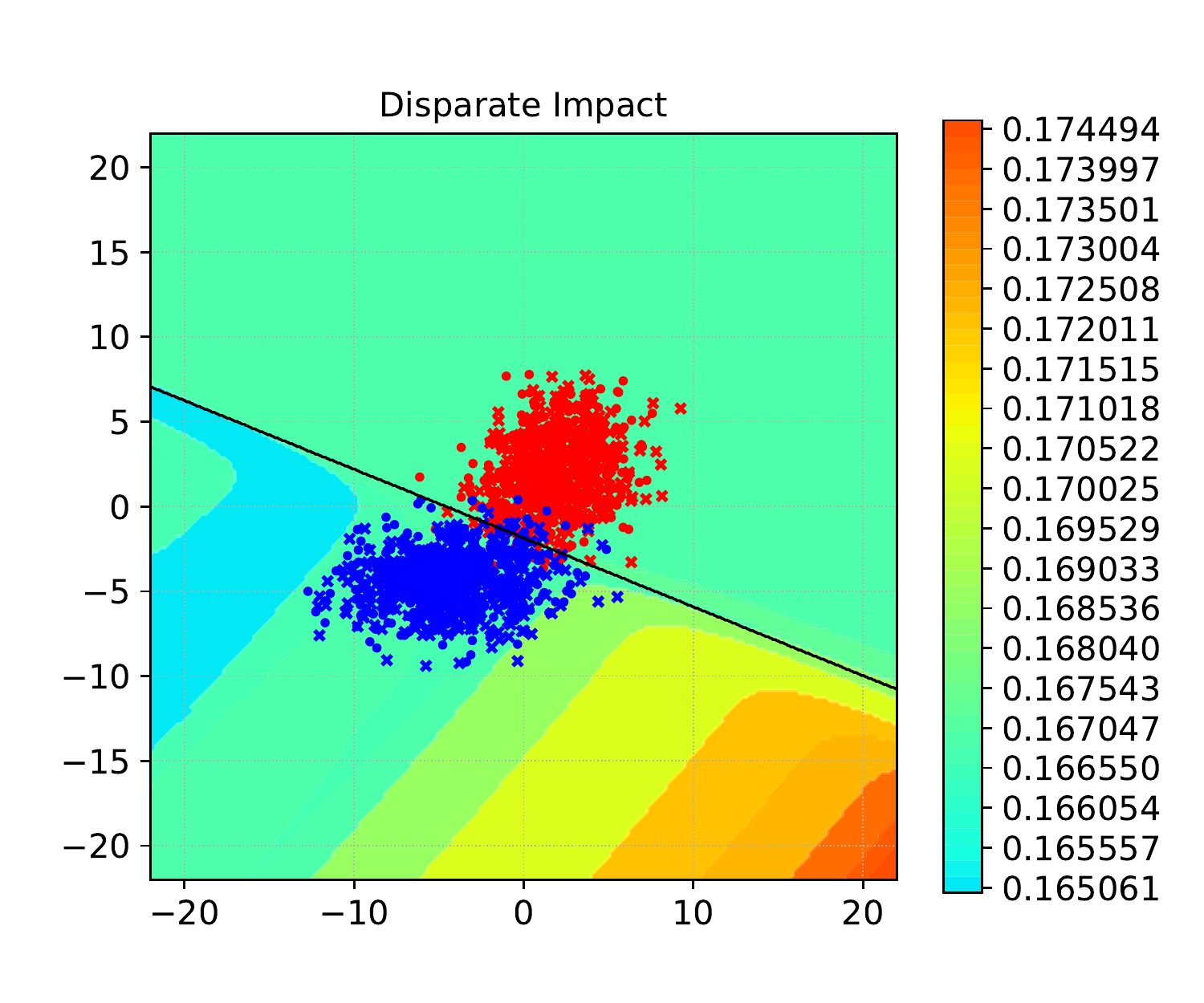}
  \caption{Attacker's loss $\set A(\vct x_c, y_c)$ (\emph{left}) and disparate impact (\emph{right}) as a function of the attack point $\vct x_c$ with $y_c=1$, on a bi-dimensional classification task. Note how the attacker's loss provides a smoother approximation of the disparate impact, and how our gradient-based attack successfully optimizes the former, which amounts to minimizing disparate impact, compromising algorithmic fairness.}
  \label{fig:poisoning_loss}
\end{figure}

In Fig.~\ref{fig:poisoning_loss}, we report a comparison of the attacker's loss $\set A(\vct x_c, y_c) = L( \set D_{\rm val}, \param^\star)$ as given by Eq.~\eqref{eq:attack_loss} and the disparate impact $D$, as a function of the attack point $\vct xc$ (with $y_c=1$) in a bi-dimensional toy example.
Each point in the plot represents the value of the function (either $\set A$ or $D$ computed on an untainted validation set) when the point $\vct x_c$ corresponding to that location is added to the training set. These plots show that our loss function provides a nice smoother approximation of the disparate impact, and that maximizing it correctly amounts to minimizing disparate impact, thus compromising algorithmic fairness.

\subsection{Gradient-Based Attack Algorithm} 

Having defining our (outer) objective, we are now in the position to discuss how to solve the given bilevel optimization problem. 
Since our objective is differentiable, we can make use of existing gradient-based strategies to tackle this problem. In particular, we will use a simple gradient ascent strategy with projection (to enforce the box constraint of Eq.~\ref{eq:poisoning_problem_box}). 
The complete algorithm is given as Algorithm~\ref{alg:poisoning}. In Fig.~\ref{fig:poisoning_loss} we also report an example of how this algorithm is able to find a poisoning point that maximizes the attacker's loss.

\emph{Attack Initialization.} An important remark to be made here is that \emph{initialization} of the poisoning samples plays a key role. In particular, if we initialize the attack point as a point which is correctly classified by the algorithm, the attack will not even probably start at all. This is clear if one looks at Fig.~\ref{fig:poisoning_loss}, where we consider an attack point labeled as positive (red). 
If we had initialized the point in the top-right area of the figure, where positive (red) points are correctly classified, the point would have not even moved from its initial location, as the gradient in that region is essentially zero (the value of the objective is constant). Hence, for a poisoning attack to be optimized properly, a recommended strategy is to initialize points by sampling from the available set at random, but then flipping their label. This reduces the risk of starting from a flat region with null gradients~\cite{biggio12-icml,biggio15-icml}. 

\begin{algorithm}[t]
	\caption{Gradient-based poisoning attack}
	\label{alg:poisoning}
	\begin{algorithmic}[1]
		\Require $\vct x_c, y_c$: the initial location of the poisoning sample and its label; $\eta$: the gradient step size; $t > 0$: a small number.
		\Ensure $\vct x_c^{\prime}$: the optimized poisoning sample.
		\State Initialize the attack sample: $\vct x_c^{\prime} \gets \vct x_c$
		\Repeat
		\State Store attack from previous iteration: $\vct x_c \gets \vct x_c^{\prime}$
		\State Update step: $ \vct x_c^{\prime} \gets \Pi \left ( \vct x_c + \eta \nabla_{\vct x_c} \set A \right ) $, where $\Pi$ ensures projection onto the feasible domain (\ie, the box constraint in Eq.~\ref{eq:poisoning_problem_box}).
		\Until{$ | \set A (\vct x_c^{\prime}, y_c) - \set A (\vct x_c, y_c) | \le t $}
		\State \Return $\vct x_c^{\prime}$
	\end{algorithmic}
\end{algorithm}

\myparagraph{Gradient Computation.} Despite the simplicity of the given projected gradient-ascent algorithm, the computation of the poisoning gradient $\nabla_{\vct x_c} \set A$ is more complicated.
In particular, we do not only need the outer objective to be sufficiently smooth w.r.t. the classification function, but also the solution $\vct \theta^\star$ of the inner optimization to vary smoothly with respect to $\vct x_c$~\cite{biggio12-icml,biggio17-aisec,demontis19-usenix,biggio18}. In general, we need $\set A$ to be sufficiently smooth w.r.t. $\vct x_c$.

Under this assumption, the gradient can be obtained as follows. First, we derive the objective function w.r.t. $\vct x_c$ using the chain rule~\cite{biggio12-icml,biggio15-icml,biggio17-aisec,biggio18,mei15-aaai}:
\begin{eqnarray}
\label{eq:poisoning_gradient}	
\nabla_{\vct{x}_c} \set{A} =  { \nabla_{\vct{x}_c} L } + {\frac{\partial \vct{\param}^\star}{\partial \vct{x}_c}}^\T {\nabla_{\vct{\param}} L} \, ,
\end{eqnarray}
where the term $\frac{\partial \vct{\param}^\star}{\partial \vct{x}_c}$ captures the implicit dependency of the parameters $\param$ on the poisoning point $\vct x$, and $\nabla_{\vct{x}_c} L$ is the explicit derivative of the outer validation loss w.r.t. $\vct x_c$. Typically, this is zero if $\vct x_c$ is not directly involved in the computation of the classification function $f$, \eg, if a linear classifier is used (for which $f(\vct x) = \vct w^\T \vct x + b$). In the case of kernelized SVMs, instead, there is also an explicit dependency of $L$ on $\vct x_c$, since it appears in the computation of the classification function $f$ when it joins the set of its support vectors~(see, \eg, \cite{biggio12-icml,demontis19-usenix}).

Under regularity of $\vct \param^\star(\vct x_c)$, the derivative $\frac{\partial \vct{\param}^\star}{\partial \vct{x}_c}$ can be computed by replacing the inner optimization problem in Eq.~\eqref{eq:poisoning_problem_inner} with its equilibrium (Karush-Kuhn-Tucker, KKT) conditions, \ie, with the implicit equation $\nabla_{\param} \set{L} (\set{D}_{\rm tr} \cup (\vct x_c, y_c ), \param) \in \vct 0$~\cite{mei15-aaai,biggio17-aisec}. By deriving this expression w.r.t. $\vct{x}_c$, we get a linear system of equations, expressed in matrix form as $\nabla_{\vct{x}_c} \nabla_{\param} \set {L}  + \frac{\partial \param^\star}{\partial \vct{x}}^\T \nabla_{\vct{w}}^2 \set{L} \in \vct 0$.
We can now compute $\frac{\partial \param^\star}{\partial \vct{x}_c}$ from these equations, and substitute the result in Eq.~\eqref{eq:poisoning_gradient}, obtaining the required gradient:
\begin{eqnarray}
\label{eq:poisoning_gradient_expanded}	
\nabla_{\vct x_c} \set{A}  = \nabla_{\vct{x_c}} L  - (\nabla_{\vct{x}_c} \nabla_{\param} \set{L} )
(\nabla_{\param}^2 \set{L})^{-1}
\nabla_{\param} L \, .
\end{eqnarray}

These gradients can be computed for various classifiers (see, e.g., \cite{demontis19-usenix}).
%
In our case, we simply need to compute the term $\nabla_{\param} L$, to account for the specific validation loss that we use to compromise algorithmic fairness (Eq.~\ref{eq:attack_loss}). 

\begin{figure}[t]
  \centering
      \includegraphics[width=0.45\textwidth]{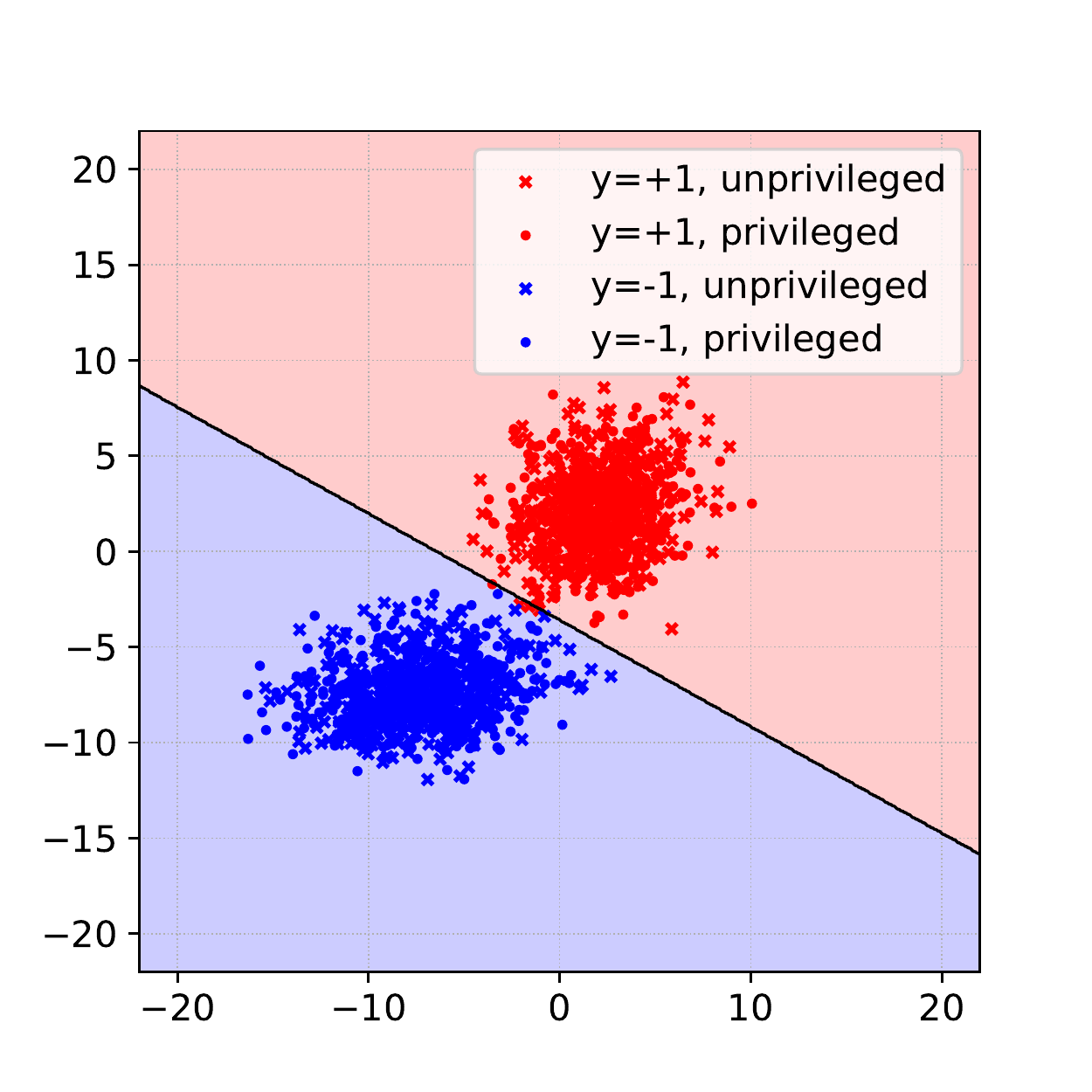}
      \includegraphics[width=0.45\textwidth]{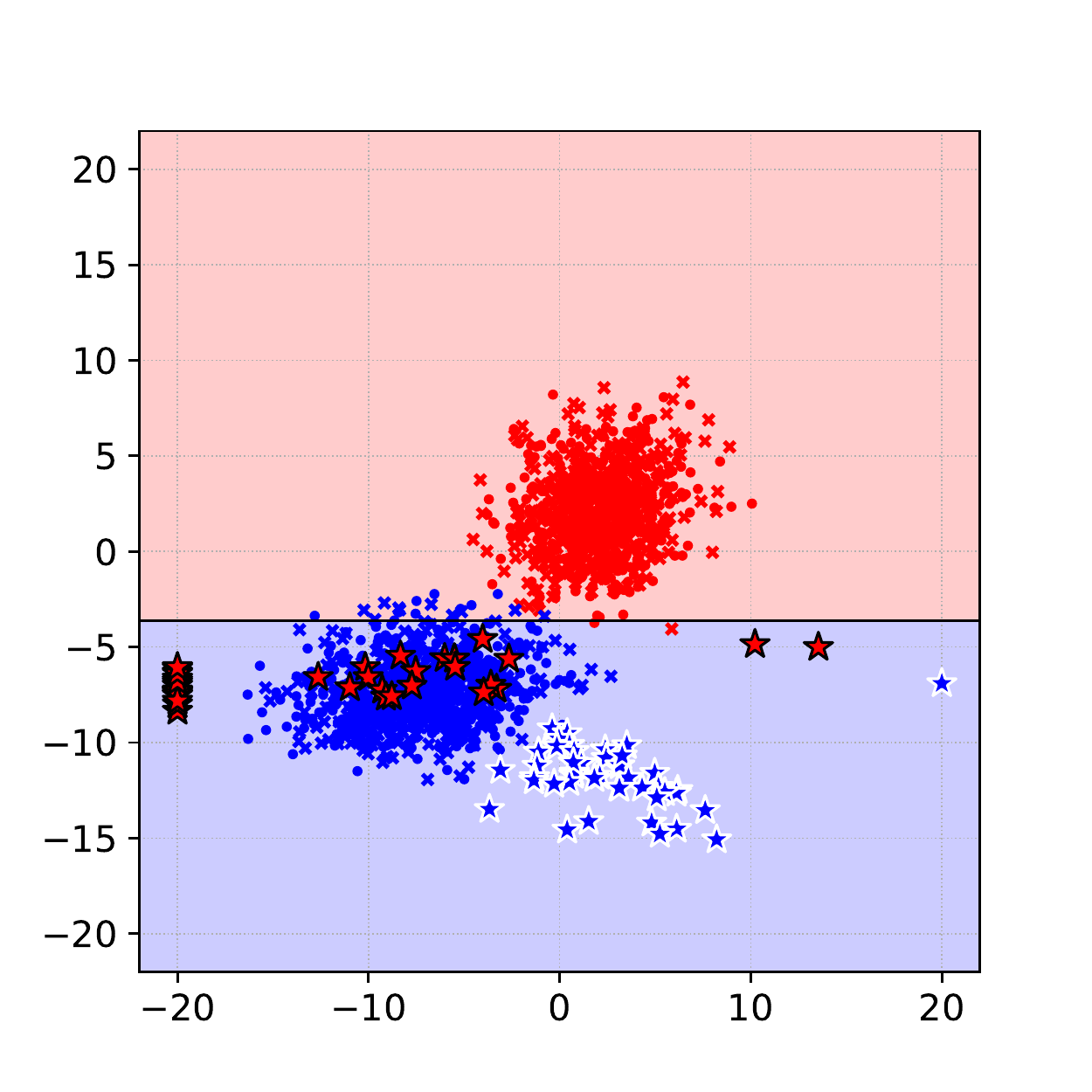}
  \caption{Gradient-based poisoning attack against a logistic classifier, on a bi-dimensional classification task. The classification function and the corresponding decision regions are reported before (\emph{left}) and after (\emph{right}) injection of the poisoning samples (red and blue stars in the right plot).}
  \label{fig:poisoning_dr}
\end{figure}

Finally, in Fig.~\ref{fig:poisoning_dr}, we show how our poisoning attack modifies the decision function of a linear classifier to worsen algorithmic fairness on a simple bi-dimensional example. As one may appreciate, the boundary is slightly tilted, causing more unprivileged samples to be classified as negative, and more privileged samples to be classified as positive.

\subsection{White-Box and Black-Box Poisoning Attacks}
\label{sec:pois_scenario}

The attack derivation and implementation discussed throughout this section implicitly assumes that the attacker has full knowledge of the attacked system, including the training data, the feature representation, and the learning and classification algorithms.
This sort of \emph{white-box} access to the targeted system is indeed required to compute the poisoning gradients correctly and run the poisoning attack~\cite{biggio18}.
It is however possible to also craft \emph{black-box} attacks against different classifiers by using essentially the same algorithm.
To this end, one needs to craft the attacks against a \emph{surrogate model}, and then check if these attack samples \emph{transfer} successfully to the actual target model.
Interestingly, in many cases these black-box transfer attacks have been shown to work effectively, provided that the surrogate model is sufficiently similar to the target ones~\cite{papernot16-transf,demontis19-usenix}.
The underlying assumption here is that it is possible to train the surrogate model on samples drawn from the same distribution as those used by the target model, or that sufficient queries can be sent to the target model to reconstruct its behavior.

In our experiments we consider both white-box attacks and black-box transfer attacks to also evaluate the threat of poisoning fairness against weaker attackers that only possess limited  knowledge of the target model.
For black-box attacks, in particular, we assume that the attacker trains the substitute models on a training set sampled from the same distribution as that of the target models, but no queries are sent to the target classifiers while optimizing the attack.

%% file: 03-experiments.tex
\section{Experiments}\label{sec:experiments}

This section describes the obtained results for two different datasets, one synthetic set composed of 2000 samples, each of them having three features, one of them considered the sensitive attribute, not used for the optimization.
The second dataset corresponds to one of the most widely used by the \textit{Algorithmic Fairness} community, a criminal recidivism prediction dataset composed by more than 6000 samples, with 18 features describing each individuals.
For each dataset, we consider both the white-box and the black-box attack scenarios described in Section~\ref{sec:pois_scenario}.
%
%
\subsection{Experiments with synthetic data}

The first round of experiments uses synthetic data set to empirically test the impact of the attacks with respect to varying levels of disparity already found in the (unaltered) training data.
Data is generated using the same approach of Zafar et al. \cite{zafar2015fairness}.
Specifically, we generate 2,000 samples and assign them to binary class labels ($y=+1$ or $y=-1$) uniformly at random.
Each sample is represented by a 2-dimensional feature vector created by drawing samples from two different Gaussian distributions:
$p(x|y=+1) \sim N([2;2], [5, 1; 1, 5]$ and
$p(x|y=-1) \sim N([\mu_1;\mu_2], [10, 1; 1,3])$
where $\mu_1, \mu_2$ are used to modify the euclidean distance $S$ between the centroids of the distributions for the privileged and unprivileged groups so that different base rates \cite{alex2016fair} can be tested in the experiments.
Then, a sample's sensitive attribute $z$ is assigned by drawing from a Bernoulli distribution using $p(z=+1) = \frac{p(x'|y=+1)}{p(x'|y=+1) + p(x'|y=-1)}$
where $x' = [cos(\phi) - sin(\phi); sin(\phi), cos(\phi)]x $ corresponds to a rotated version of the feature vector $x$.

\begin{figure}[t]
  \centering
      \includegraphics[width=1\textwidth]{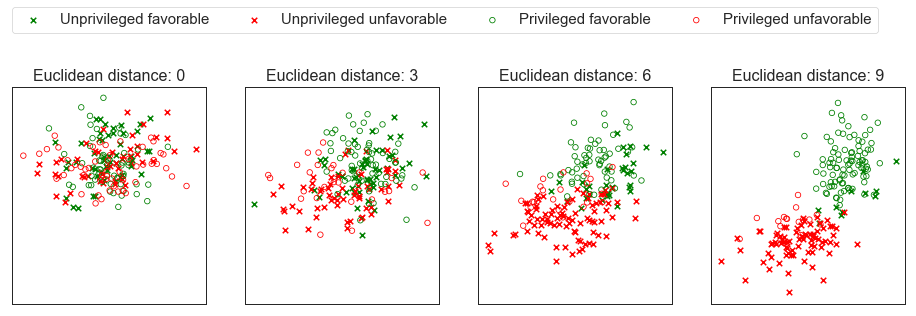}
  \caption{(Best seen in color.) Examples of generated synthetic data sets for different values of the separation $S$ between groups.%
  Privileged elements ($z=+1$) are denoted by circles and unprivileged elements ($z=-1$) by crosses.
  Favorable labels ($y=+1$) are in green, while unfavorable labels ($y=-1)$ are in red.}
  \label{fig:Synthetic_dataset}
\end{figure}

Using the generator we have described, datasets such as the ones as depicted in Figure \ref{fig:Synthetic_dataset} can be obtained. 
In this figure, the feature vector $x$ is represented in the horizontal and vertical axes, while
the color represents the assigned label $y$ (green means favorable, red means unfavorable) and
the symbol the sensitive attribute $z$ (circle means privileged, cross means unprivileged).

We generate multiple datasets by setting $S \in \{0, 1, 2, \dots, 9\}$.
We then split each dataset into training $D_{tr}$ (50\% of the samples), validation $D_{val}$ (30\%) and testing $D_{test}$ (20\%) subsets. 
In each run, a base or initial model $\set M$ is trained.
This model $\set M$ corresponds to a Logistic Regression model in the first setting and to a Support Vector Machine with linear kernel in the second scenario.
The regularization parameter $C$ is automatically selected between $[0.5, 1, 5, 10]$ through cross validation.
In the \textit{White-Box} setting, the attack is optimized for $\set M$ so that Eq. \ref{eq:obj-pois} is minimized in the training set $D_{tr}$ and Eq. \ref{eq:poisoning_problem_box} is maximized in the validation set $D_{val}$.
In the \textit{Black-Box} setting, the attack is optimized against a surrogate model $\hat{\set M}$, a Logistic Regression classifier, trained with another subset of data generated for the same value of the parameter $S$ 
Each of these attacks generates a number of poisoning samples.
The poisoned model is the result of retraining the original model with a training set that is the union of $D_{tr}$ and the poisoned samples. 


The attack performance is measured by comparing the model trained on the original training data with a model trained on the poisoned data.
The evaluation is done according to the following metrics, which for each dataset are averaged over ten runs of each attack:

\begin{itemize}
\item \textbf{Accuracy} The accuracy on test obtained by the poisoned model is similar and correlated with the accuracy obtained by a model trained on the original data.
It is important to note that the separability of the generated data is also highly correlated with the separation between the groups in the data, creating this effect.
%
\item \textbf{Demographic parity} Measures the allocation of positive and negative classes across the population groups. Framed within the Disparate impact criteria that aims to equalize assigned outcomes across groups, this metric is formulated as: 
$$
P(\hat Y=1 | G= unprivileged) - P(\hat Y=1 | G= privileged)
$$
It tends to zero in a fair scenario and is bounded between $[1,-1]$ being -1 the most unfair setting. This metric is correlated with the \textit{Disparate impact} metric introduced in Section \ref{sec:algorithm} and has been selected for convenience in the visual representation of the results.
\item \textbf{Average odds difference} The average odds difference is a metric of disparate mistreatment, that attempts for Equalized odds \cite{fair_classification_6}, it accounts for differences in the performance of the model across groups. This metric is formulated as:
$$ \frac{1}{2}[(FPR_{p} - FPR_{u}) + (TPR_{p} - TPR_{u})]
$$
It gets value zero in a fair scenario and is bounded between $[1,-1]$ being -1 the most unfair setting.

\item {\textbf{FNR privileged} False Negative Rate for the privileged group of samples.}

\item {\textbf{FNR unprivileged} False Negative Rate for the unprivileged group of samples.}

\item {\textbf{FPR privileged} False Positive Rate for the unprivileged group of samples.}

\item {\textbf{FPR unprivileged} False Positive Rate for the unprivileged group of samples.}
\end{itemize}

\begin{figure}[htb]
        \centering
        \includegraphics[width=1
        \textwidth]{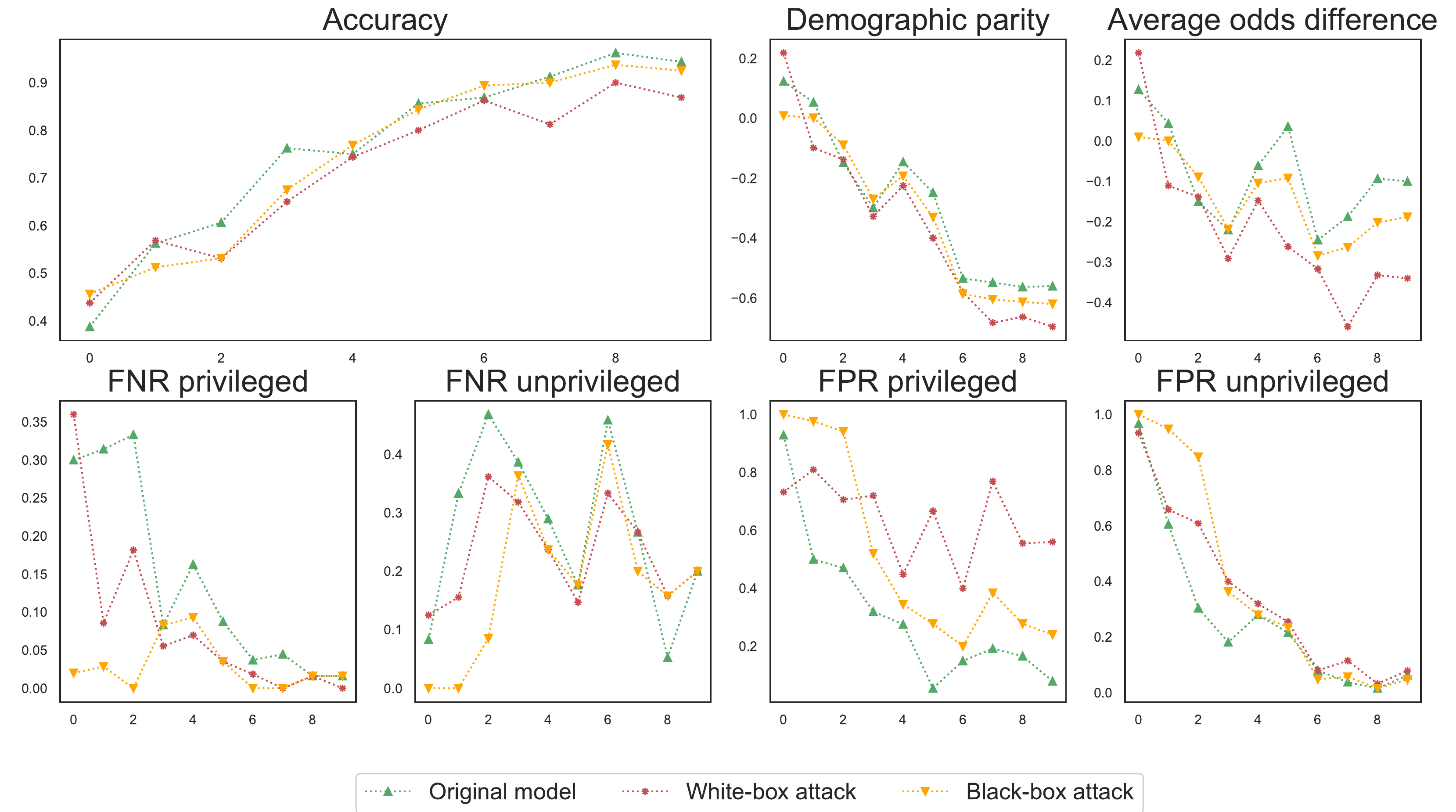}
    \caption{Comparison of the original model against the model generated by the White-box attack and Black-box attacks, for ten synthetic datasets generated by different separation parameters ($S$).
    Each data point is the average of ten runs of an attack.
    We observe that attacks have a moderate effect on the accuracy of the classifier, and can affect the classifier fairness (demographic parity and odds difference) to an extent that becomes more pronounced if the original dataset already has a large separation between classes (larger values of $S$).}
    \label{fig:dimp_attack_white}
\end{figure}

Results shown on Figure~\ref{fig:dimp_attack_white} show the obtained performance of the attacks for the generated data.
In this figure, the horizontal axis is the separation $S$ between classes in each of the ten datasets.
Analyzing the results, we observe that the poisoned models increase disparities in comparison with a model created on the unaltered input data, across all settings.
Additionally, they yield
an increased FPR for the privileged group (privileged samples that actually have an unfavorable outcome are predicted as having a favorable one), increasing significantly the observed unfairness as measured by the fairness measurements.
We note that the attacks also decrease the FNR of the unprivileged group (unprivileged samples that actually have a favorable outcome are predicted as having an unfavorable one).
This is most likely a consequence of the attack's objective of maintaining accuracy and show that this attack is not trivial. If the attack were only to increase disparities, it would also increase the FNR of the unprivileged group with a larger decrease in accuracy than what we observe.
The decrease of FNR for the unprivileged group, however, is smaller than the increase of FPR for the privileged group, as the average odds difference plot shows, and hence the attack succeeds.

\subsection{Experiments with real data}

To demonstrate the attacks on real data, we use the COMPAS dataset released by ProPublica researchers~\cite{ProPublica_article}, which is commonly used by researchers on Algorithmic Fairness.
This dataset contains a prediction of criminal recidivism based on a series of attributes for a sample of $6,167$ offenders in prison in Broward County, Florida, in the US.
The attributes for each inmate include criminal history features such as the number of juvenile felonies and the charge degree of the current arrest, along with sensitive attributes: race and gender.
%
%
For each individual, the outcome label (``recidivism'')  is a binary variable indicating whether he or she was rearrested for a new crime within two years of being released from jail.


We use this dataset for two different types of experiments.
First, we show how the attacks demonstrated on synthetic data can also be applied to this data,
and demonstrate the effect of varying the amount of poisoned samples,
Second, we evaluate the transferability of the attack to other classification models.

\myparagraph{White-Box and Black-Box poisoning attacks with varying amounts of poisoned samples.}
This experiment compares the original model against the model obtained under the two attack models. 

\begin{figure}[H]

        \centering
        \begin{subfigure}{1\textwidth}
        \includegraphics[width=1
        \textwidth]{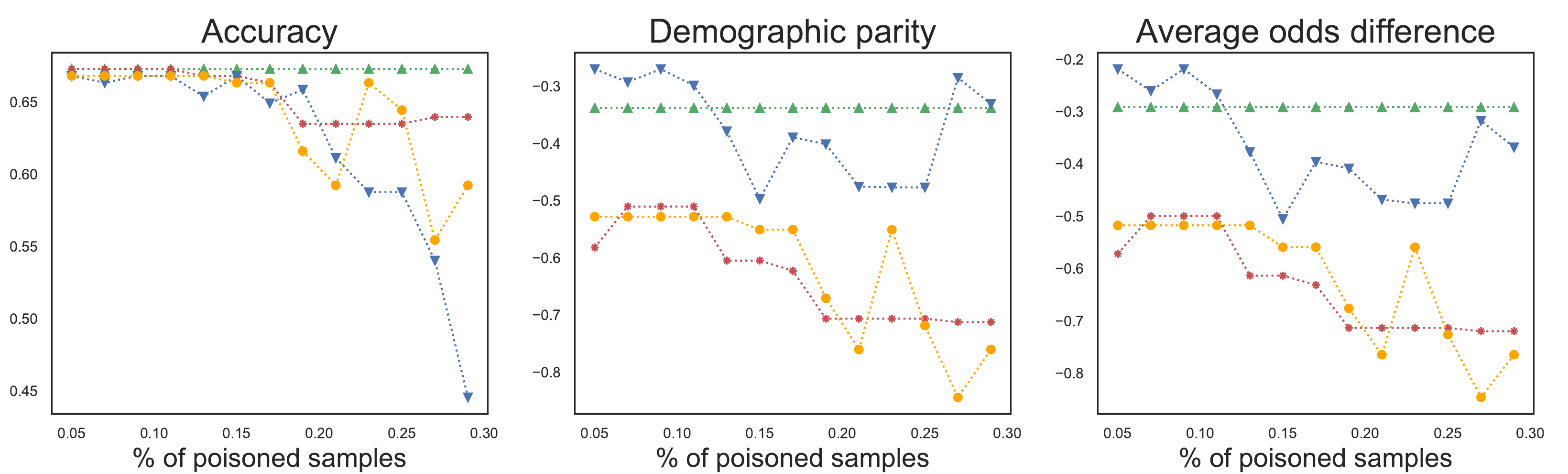}
        \end{subfigure}\hfil
        
        \begin{subfigure}{1\textwidth}
         \centering
        \includegraphics[width=1
        \textwidth]{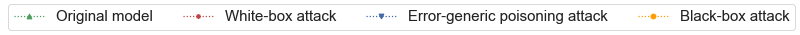}
        \end{subfigure}\hfil
        
    \caption{Comparison of the original model against the model generated by a White-box attack and a Black-box attack, for varying percentages of poisoned samples.
    The main difference between both types of attack is that the black-box attack starts having more noisy behaviour also drastically reducing the accuracy of the classifier (thus being more easily detectable) when the percentage of poisoned samples exceeds a certain threshold (about 20\%).}
    \label{fig:attack_compas_increment}
\end{figure}

Figure \ref{fig:attack_compas_increment} shows the results, which are in line with the findings of the experiments on synthetic data.
According to the obtained results, both types of poisoning attacks are is able to increase unfairness of the model with a more modest effect on the accuracy. Also, an interesting finding is the stability of the \textit{White-Box} attack as opposite to the \textit{Black-Box} attack. Whereas the first keeps the same trend with the growing number of samples, the later starts having a unstable and noisy behaviour after adding the 20\% of samples, causing for some cases a more unfair model but also affecting the accuracy of the system in a manner that could be easily detected.

In Figure \ref{fig:attack_compas_increment} we also include an Error-Generic Poisoning Attack \cite{demontis19-usenix} for the Logistic Regression model , which is designed to decrease the accuracy of the resulting model.
We observe that this type of generic adversarial machine learning attack does not affect the fairness of the classifier nearly as much as the attacks we have described on this paper.

As expected, computing the obtained performance for all the stated metrics, (Figure omitted for brevity) can be observed that the effect of any attack increases with the number of poisoned samples.
In general, these attacks
increase the False Negatives Rate (FNR) for the unprivileged samples, and
increase the False Positives Rate (FPR) for the privileged samples.

\myparagraph{Transferability of the attack.}
We study how an attack would affect the performance of other type of models, simulating different scenarios of \textit{Zero Knowledge} attacks.

Specifically, the attacks we perform is optimized for a Logistic Regression model, and its performance is tested for other models:
\begin{inparaenum}[(a)]
\item Gaussian Naive Bayes.
\item Decision Tree; 
\item Random Forest; 
\item Support Vector Machine with linear kernel; and
\item Support Vector Machine with Radial Basis Function (RBF) kernel.
\end{inparaenum}

\begin{figure}[H]
        \centering
        \begin{subfigure}{1\textwidth}
        \includegraphics[width=1
        \textwidth]{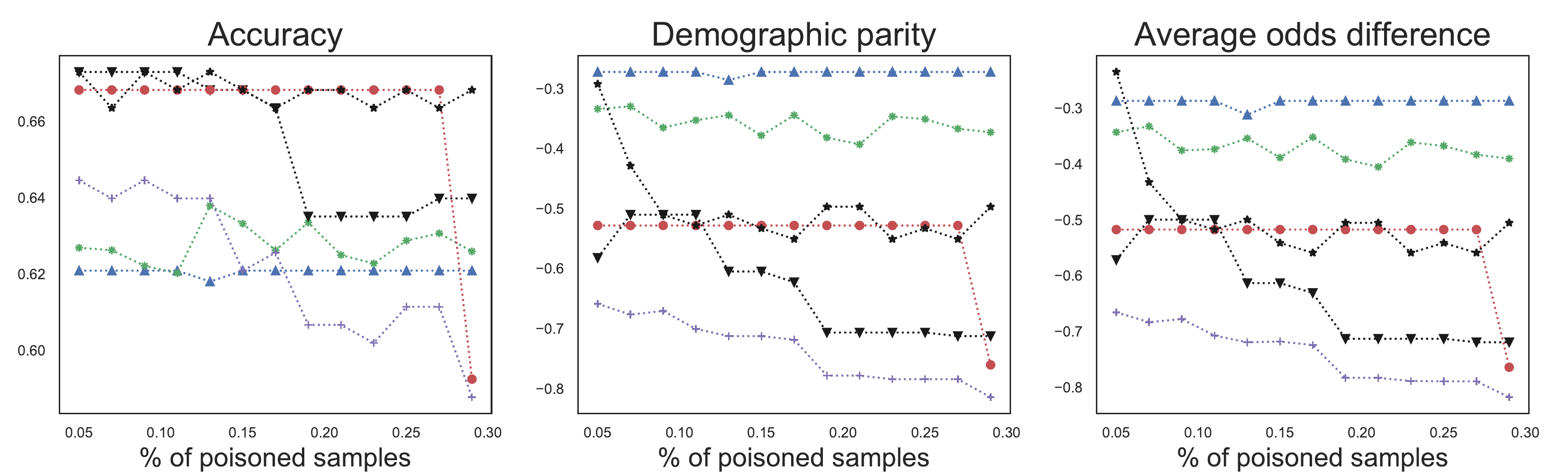}
        \end{subfigure}\hfil
        
        \begin{subfigure}{1\textwidth}
         \centering
        \includegraphics[width=0.6
        \textwidth]{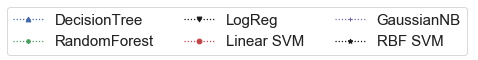}
        \end{subfigure}\hfil
        
    \caption{Transferability of the attacks from Logistic Regression to other models.}
    \label{fig:attack_compas_transf}
\end{figure}

Results are shown on Figure \ref{fig:attack_compas_transf}, in which each data point corresponds to the average of five experimental runs.
We observe that the attack optimized on a Logistic Regression classifier has a stronger effect on the Logistic Regression, Support Vector Machine (for both types of kernel tested) and Naive Bayes models.
In contrast, while it can introduce unfairness through demographic disparity and average odds difference on a Decision Tree or Random Forest classifier, its effects are more limited.

%% file: 04-related_work.tex
\section{Related Work}\label{sec:related-work}

\myparagraph{Adversarial Machine Learning Attacks.}
This work is based on Gradient-Based Optimization, an optimization framework widely used in the literature on Adversarial Machine Learning for crafting poisoning attacks \cite{biggio12-icml,mei15-aaai,biggio17-aisec,jagielski2018manipulating,demontis19-usenix}.
Such framework is used to solve the bilevel optimization given by Eqs.~\eqref{eq:obj-pois}-\eqref{eq:poisoning_problem_box}, and requires computing the gradient of the classification function learned by the classifier. As a result, poisoning samples can be obtained by iteratively optimizing one attack point at a time \cite{biggio15-icml}.

\myparagraph{Measuring Algorithmic Fairness.}
Many different ways of measuring algorithmic fairness have been proposed~\cite{narayanan2018translation}.
Among those that can be applied in an automatic classification context we find two main types: individual fairness metrics and group fairness metrics~\cite{hajian2016algorithmic}.
The former seek \emph{consistency} in the sense that similar elements should be assigned similar labels~\cite{dwork2012fairness}.
The latter seek some form of \emph{parity}, and in many cases can be computed from a contingency table indicating the number of privileged and unprivileged samples receiving a positive or negative outcome~\cite{Pedreschi_2012}.
Popular group fairness metrics include disparate impact, equalized odds~\cite{fair_classification_6}, and disparate mistreatment~\cite{Zafar_2017}.

\myparagraph{Optimization-Based Approaches to Increase Fairness.}
Algorithmic fairness can and often is compromised unintentionally, as discrimination in machine learning is often the result of training data reflecting discriminatory practices that may not be apparent initially~\cite{Barocas_NIPS_2017}.
When this is the case, training data can be modified by a type of poisoning attack, in which so-called ``antidote'' samples are added to a training set to reduce some measure of unfairness.
One such approach proposes a method to be applied on recommender systems based on matrix factorization \cite{Rastegarpanah_2019}; another is based in the Gradient-Based Optimization framework used in this work \cite{Kulynych_2020}.
%


In addition to methods to mitigate unfairness by modifying training data (something known as a pre-processing method for algorithmic fairness~\cite{hajian2016algorithmic}), other methods modify the learning algorithm itself to create, for instance, a fair classifier~\cite{zafar2015fairness,Zafar_2017} 
In these works, the trade-off between accuracy and fairness is approached through an alternative definition of fairness based in covariance between the users’ sensitive attributes and the signed distance between the feature vectors of misclassified users and the classifier decision
boundary.

%% file: 05-conclusions.tex
\section{Conclusions and Future Work}\label{sec:conclusions}

The results show the feasibility of a new kind of adversarial attack crafted with the objective of increasing disparate impact and disparate mistreatment at the level of the system predictions. 
We have demonstrated an attacker effectively alter the algorithmic fairness properties of a model even if pre-existing disparities are present in the training data.
This means that these attacks can be used to both introduce algorithmic unfairness, as well as for increasing it where it already exists.
This can be done even without access to the specific model being used, as a surrogate model can be used to mount a black-box transfer attack.

%
Studying adversarial attacks on algorithmic fairness can help to make machine learning systems more robust. 
Additional type of models such neural nets and/or other data sets can be considered in the future to extend the work proposed here.
Although experiments in this paper are done using a specific technique based on a poisoning attack, other techniques can be certainly considered.
Other approaches such as causality-based techniques could be explored as future work. 

%% file: 06-acknowledgements.tex
\section{Acknowledgements}
This research was supported by the European Commission through the ALOHA-H2020 project. Also, we wish to acknowledge the usefulness of the Sec-ML library \cite{melis2019secml} for the execution of the experiments of this paper.
C. Castillo thanks La Caixa project LCF/PR/PR16/11110009 for partial support.
B. Biggio acknowledges that this work has been partly funded by BMK, BMDW, and the Province of Upper Austria in the frame of the COMET Programme managed by FFG in the COMET Module S3AI.